\documentclass[letterpaper]{article} 
\usepackage{aaai2026}  
\usepackage{times}  
\usepackage{helvet}  
\usepackage{courier}  
\usepackage[hyphens]{url}  
\usepackage{graphicx} 
\usepackage{natbib}  
\usepackage{caption} 
\usepackage{algorithm}
\usepackage{algorithmic}
\usepackage{amsmath}

\usepackage{newfloat}
\usepackage{listings}
\DeclareCaptionStyle{ruled}{labelfont=normalfont,labelsep=colon,strut=off} 
\floatstyle{ruled}
\newfloat{listing}{tb}{lst}{}
\floatname{listing}{Listing}
\usepackage{multirow}

\title{%
  \raisebox{-0.2\height}{%
    \includegraphics[height=3.5em,keepaspectratio]{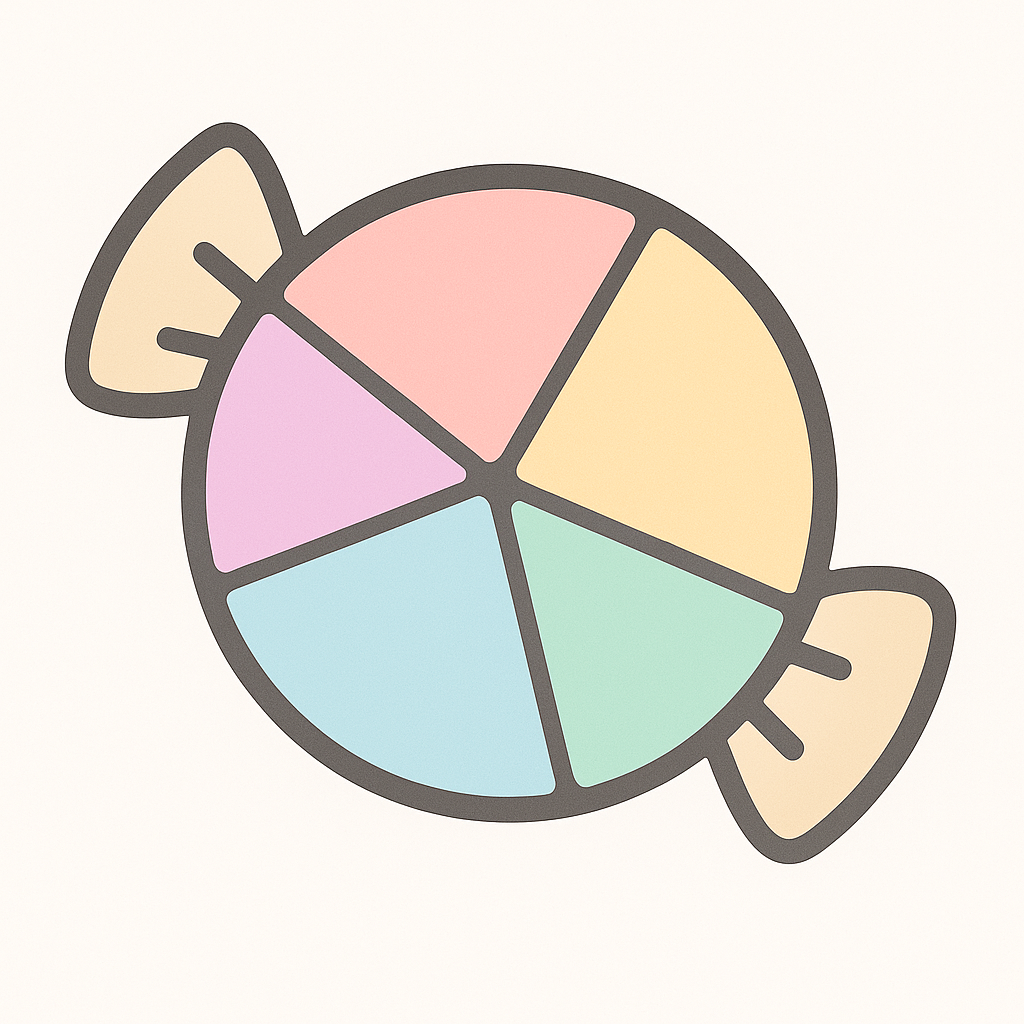}%
  }%
  CANDI – Contextual Alignment for Niche Domains Question Answering
}
\author{
    Megha Chakraborty\textsuperscript{\rm 1}\thanks{Preprint. Under review. For correspondence: megha.chakraborty@iairo.ai},
    Darssan L. Eswaramoorthi\textsuperscript{\rm 1},
    Het Riteshkumar Shah\textsuperscript{\rm 2},
    Madhur Thareja\textsuperscript{\rm 3},
    Michelle A Ihetu\textsuperscript{\rm 1},
    Harshul Raj Surana\textsuperscript{\rm 1},
    Kaushik Roy\textsuperscript{\rm 1},
    Amit Sheth\textsuperscript{\rm 1}
}

\affiliations{
    \textsuperscript{\rm 1}University of South Carolina\\
    \textsuperscript{\rm 2}IIIT-Delhi\\
    \textsuperscript{\rm 3}Indian Institute of Technology Madras\\
}

\usepackage{bibentry}

\begin{document}

\maketitle

\begin{abstract}
The increasing deployment of large language models (LLMs) in specialized, real-world domains like medical diagnostics and financial advisory demands a critical re-evaluation of their capabilities beyond general knowledge. Traditional question answering benchmarks offer limited insight into how well models adapt to the nuanced contextual grounding, user awareness, and domain understanding required in these fields. This gap highlights a pressing need for evaluation methodologies that reflect the complexities of niche-domain interactions.

To address this, we introduce \textbf{CANDI-QA} (Contextual Alignment for Niche Domains Question Answering), a novel dataset designed to evaluate LLMs on their ability to provide accurate, context-sensitive, and user-aligned answers in specialized settings. CANDI-QA comprises question-answer pairs derived from expert-curated domain-specific documents and case scenarios, structured into two distinct categories:

\begin{itemize}
    \item \textbf{T1: Information Assistance Questions-} Direct, factual queries requiring precise information extraction.
    \item \textbf{T2: Applied Inference Questions-} Complex, multi-hop reasoning questions necessitating inference based on role-specific or situational cues to generate actionable insights.
\end{itemize}

We conduct systematic evaluations across over ten diverse language models, from compact open-source systems to state-of-the-art proprietary systems. To establish a robust baseline, we present a lightweight neuro-symbolic solution named the \textit{\textbf{MTSS-Net framework}}, which combines neural network-based retrieval with rule-based reasoning.

Our findings underscore the profound challenges of achieving true contextual alignment in niche domains and highlight the limitations of current LLMs in fully supporting user-aligned, real-world applications without enhanced contextual integration or symbolic reasoning.

\textbf{CANDI-QA} serves as a critical new benchmark for advancing research in context-aware and user-aligned language models, stimulating further exploration into more robust and trustworthy AI systems for high-stakes domains.
\end{abstract}


\section{Introduction}
\label{sec:Introduction}


In 2014, Stuart Russell introduced the ``Value Alignment Problem" in an interview \cite{myth_ai},  emphasizing the need to construct AI
systems that are not just intelligent but also aligned with human values. Aligning LLM responses to user preferences has been widely studied since then. The alignment problems specifically identify the dangers of generating harmful, inaccurate, or unsafe responses.

While most bodies of work on the alignment problem mostly talk about Artificial General Intelligence tasks, we find that real-world scenarios or enterprise applications necessitate the need for contextual alignment, specifically when the users are multi-disciplinary stakeholders.

We define \textbf{\textit{contextual alignment}} as a model’s ability to align its responses to the specific context of conversation, which could imply the domain/ discipline of the user, what role they play in their domain, what their background knowledge/ domain expertise is on, or very specifically to the exact topic of conversation. 

Aligning LLM responses to the context specific to the user within the topic of conversation becomes a necessity when building a conversational model for a niche domain. We deem it necessary to augment existing datasets to provide directed context for them to be more useful in domain applications has an impact on real-world/ enterprise cases. In this paper, we illustrate the importance/ significance of contextual alignment through a very critical, real-world use case of implementing the Multi-Tier System of Supports (MTSS) framework in schools for Behavioral Health.

We evaluate the performance of \textbf{10} language models: Phi-4-mini-instruct (3.8B), Deepseek R1, Qwen3 (32B), Gemma 2 (9B), Gemini1.5, LLaMA 3.1 (8B) - Instruct, LLaMA 3.2 (1B) - Instruct, LLaMA 4 Maverick (7B), LLaMA 4 Scout (17B), and Mistral 7B Instruct. in the question-answering downstream task for a niche domain such as MTSS for Behavioral Health (MTSS(B)). \\

The key contributions of this paper are as follows:
\begin{itemize}
    \item We propose a new task called \textbf{Contextual Alignment for Niche Domains Question Answering (CANDI-QA)}.
    
    \item We present a consolidated and comprehensive dataset consisting of two types of questions:
    \begin{itemize}
        \item \textbf{T1}: Information Assistance Questions
        \item \textbf{T2}: Applied Inference Questions
    \end{itemize}
    
    \item We perform systematic evaluations on state-of-the-art (SOTA) language models for question answering and establish a benchmark to evaluate performance in the CANDI-QA setting.
    
    
    \item We provide a baseline solution leveraging neuro-symbolic AI approaches. A simple solution that is implemented through a rule-based or logic-based methodology called the \textbf{MTSS-Net} framework.
\end{itemize}

\begin{figure}[h]
    \centering
    \includegraphics[width=\columnwidth]{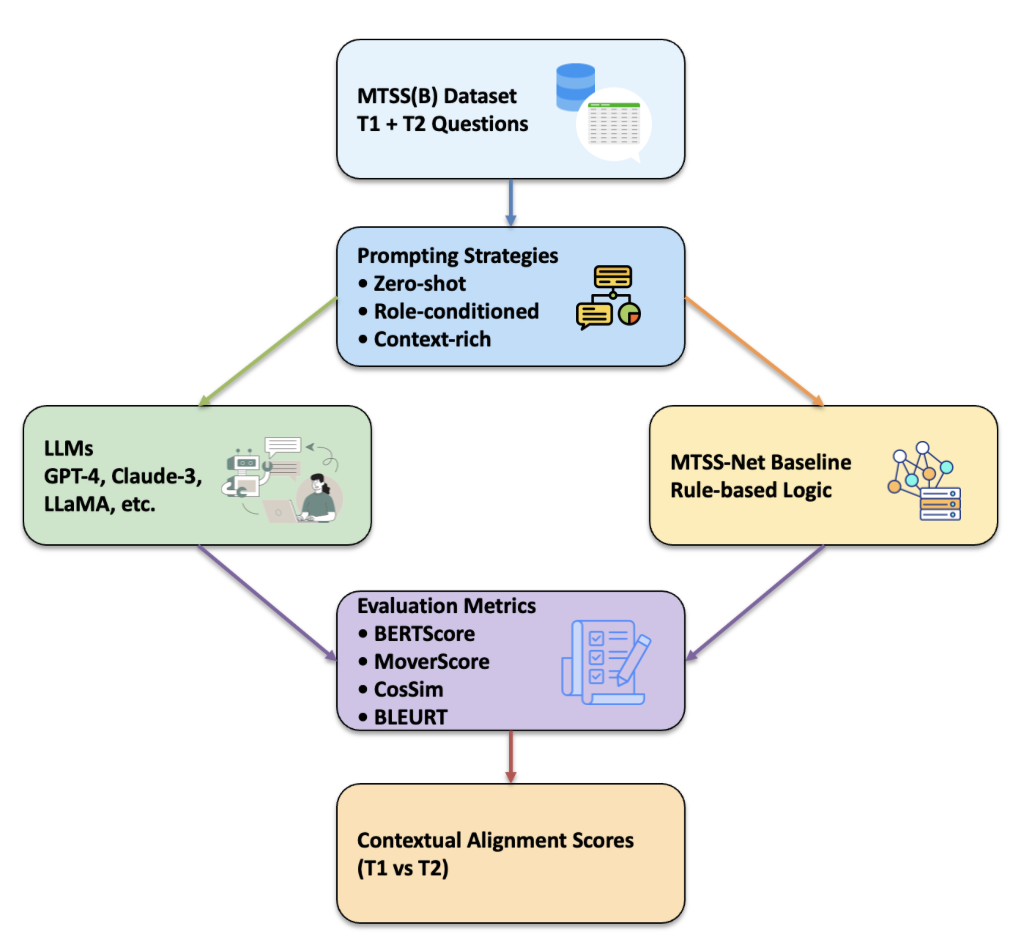}
    \caption{CANDI Architecture}
    \label{fig:singlecol}
\end{figure}
\section{Related Works}
\label{sec:RelatedWorks}


\subsection{From General to Domain-Specific Question Answering}

The evolution of Question Answering (QA) systems has progressed from open-domain architectures---where general knowledge corpora and benchmarks dominate---to increasingly specialized frameworks built for domain-specific or context-sensitive tasks. Early datasets and benchmarks such as SQuAD, Natural Questions, and TriviaQA fostered precision in retrieving and synthesizing fact-based answers from broad knowledge sources \cite{rajpurkar2016squad, joshi-etal-2017-triviaqa}. However, these resources provided limited support for scenarios requiring nuanced, role- or domain-aware comprehension.

Recent efforts have targeted the limitations of standard large language models (LLMs) when applied to specialized or niche domains. Closed-book and retrieval-augmented approaches, such as the two-stage context generation method, enhance LLM responses by conditioning on personalized or contextually generated input \cite{verma2024hybridcontextqa, yang2023empower}. These techniques have demonstrated improvements in exploiting parameterized knowledge for QA, narrowing the performance gap between general-domain and context-rich settings.

\subsection{Construction of Domain-Specific QA Datasets}

Building robust QA datasets tailored for niche domains typically demands new methodologies. Recent work emphasizes semi-automatic data acquisition, leveraging generative models for question and/or answer generation, with targeted human annotation to ensure domain fidelity and cost-effective scaling. For example, Falk et al. (2024) demonstrate that selective manual labeling combined with automated augmentation yields superior adaptation and higher QA performance compared to fully manual or fully synthetic datasets \cite{falk2024exploring}.


Notably, M-QALM and DomainCQA introduce methodologies for flexible, scalable dataset generation in complex scientific and technical domains \cite{subramanian2024m, zhong2025domaincqa}. They distinguish between fundamental, fact-based questions and advanced, inference-driven or context-dependent questions. Both utilize LLMs not only for data generation, but also for annotation and evaluation, reducing reliance on scarce domain experts and enabling research at scale.

\subsection{Annotation, Validation, and Evaluation in Niche Domains}

Ensuring quality and reliability in domain-specific QA datasets requires multi-stage validation. Manual annotation by domain experts, selective bootstrapping, and cross-dataset error analysis are commonly employed \cite{singhal2022large, racha2025mhqa}. Human-centric qualitative metrics---such as factuality, completeness, and hallucination rate---are increasingly adopted alongside traditional metrics (e.g., BLEU, ROUGE, F1) to judge alignment with expert gold standards and real-world needs.

Furthermore, rubric-based and reference-free evaluation metrics, as implemented in frameworks like RAGAS, enable nuanced judgment of contextual correctness and domain alignment \cite{es2024ragas}. These approaches are crucial in settings where literal answer matching misses contextual appropriateness or logical validity.

\subsection{Neuro-Symbolic Architectures as Baselines}

Hybrid QA models integrating LLMs with retrieval components and symbolic reasoning engines are gaining traction for their enhanced interpretability and modularity. For instance, neuro-symbolic approaches---such as NSQA and recent hierarchical paradigms---combine neural models for question understanding with logic-based components for rule-driven reasoning, supporting transparent, expert-understandable decision making \cite{kapanipathi2021leveraging, zhang2024neuro}.

\subsection{Benchmarking and Performance in Contextual Alignment}

Benchmarks now encompass both standard and domain-adapted datasets, assessing LLMs - from compact open-source variants to proprietary giants \cite{manes2024kqa, akinseloyin2024question}. Comparative studies consistently show (a) retrieval-augmented and role-aware prompting strategies outperforming vanilla LLMs, and (b) significant headroom remaining for models to achieve expert-level contextual reasoning, especially on highly-specific or inferential queries \cite{racha2025mhqa, subramanian2024m}.
\section{Problem Formulation}
\label{sec:ProblemFormulation}

\subsection{Domain - Implementing MTSS for Behavioral Health}

The core MTSS(B) is a multidisciplinary team that includes leaders at both district and school levels who coordinate to align practices, programs, and personnel to implement services at three levels of impact: 
\begin{itemize}
    \item Tier 1: School wide/ Universal level
    \item Tier 2: Targeted, small group level
    \item Tier 3: Intensive individual care
\end{itemize}

The Multi-Tiered System of Supports (MTSS) framework involves both school-level and district-level teams comprising diverse stakeholders. School-level teams typically include administrators, mental health professionals (licensed and non-licensed), teachers, counselors, nurses, resource officers, community partners, caregivers, and interns or coaches at select sites. District-level teams feature executive leaders such as superintendents, directors of student services, special education, and mental health, as well as representatives from state agencies, families, and community organizations. School teams are responsible for identifying and supporting students needing academic, behavioral, or social-emotional interventions through data-driven assessment, tiered support, progress monitoring, and collaborative problem-solving. District teams oversee policy, resource allocation, fidelity monitoring, professional development, and data analysis to support and ensure consistent MTSS implementation across schools.

Although state-mandated laws exist for a majority of US states, the effective implementation of MTSS in schools is challenging and complex because the MTSS framework involves numerous components and requires a collaborative effort from a multidisciplinary team to track tier-assignment related information.

Teams often face limitations in their awareness and capacity to implement best practices effectively, and turn to experts or coaches for support. However, the availability of such coaching support is typically limited, with most coaches operating at the district level and serving large numbers of schools—resulting in a significant gap between the recommended and actual coach-to-school ratios. This disparity underscores the potential role of AI-based tools in augmenting human capacity and providing scalable, real-time support for MTSS teams.

\begin{figure}[t]
    \centering
    \setlength{\abovecaptionskip}{-4pt}
    \includegraphics[width=\columnwidth]{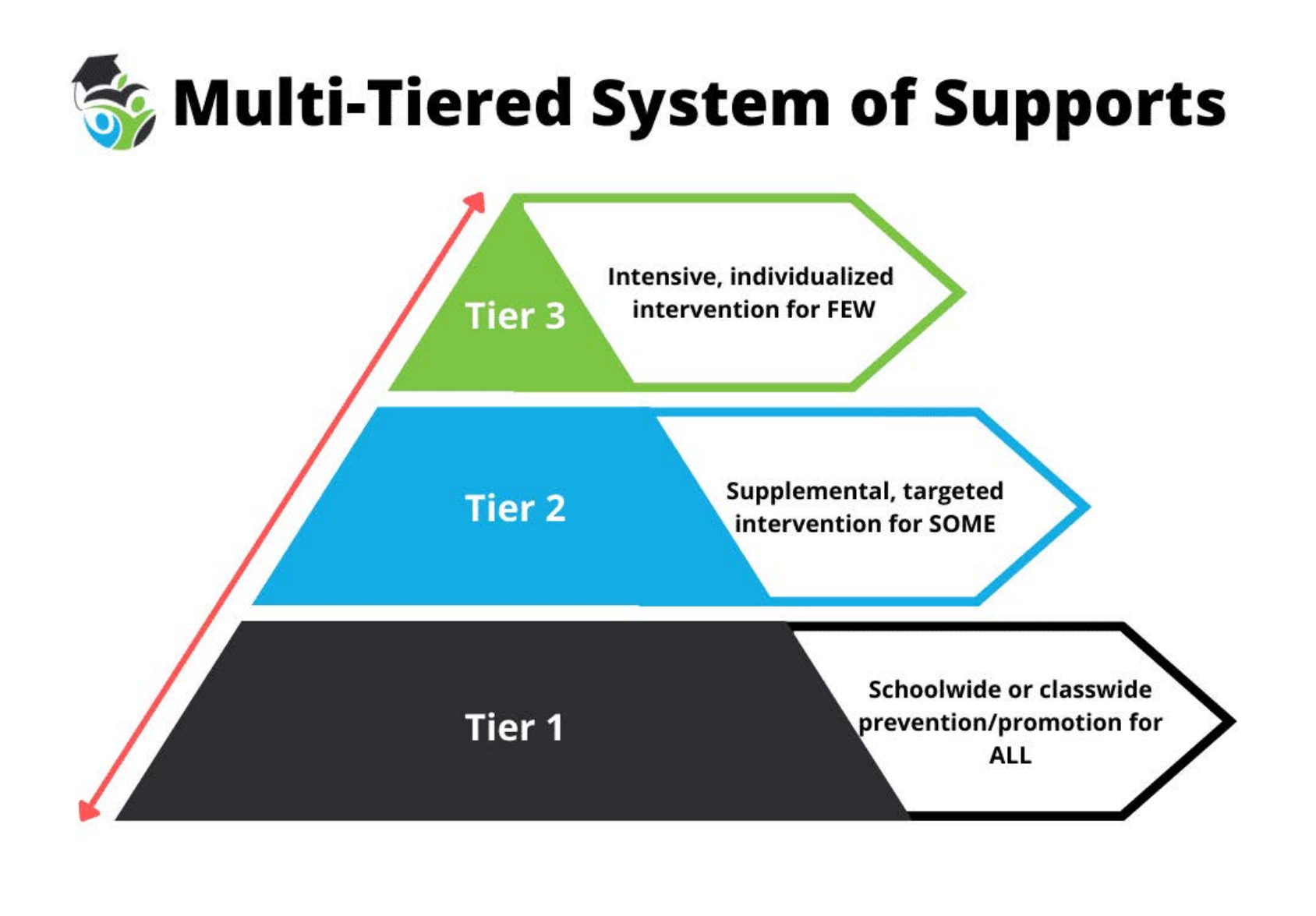}
    \caption{MTSS (B)}
    \label{fig:bottomfig}
\end{figure}

\subsection{Task Definition - CANDI-QA} 

For effective intervention, the MTSS Team usually has questions that can be categorized as 
\begin{enumerate}
    \item T1, Information assistance questions - where the answers are available and accessible through expert curated literature on best-practices, evidence-based practices, MTSS literacy etc.
    \item T2, Applied inference questions - where the answers are not directly available in the literature, but need understanding, experience, and reasoning to mitigate or respond to the pressing concerns. These questions are typically more student-specific or scenario-specific.
\end{enumerate}

Typically, these questions are addressed to MTSS Coaches for guidance and support from their knowledge and experience. Since the coaches are a limited human resource, they face too much cognitive overload to address concerns from multiple MTSS Teams working in different schools and districts. Working towards the goal of integrating AI within MTSS teams, we design our model objective as follows: \\
 `You are an AI assistant for the multidisciplinary team members implementing MTSS for Behavioral health in schools. Your objective is to assist them with data-driven, decision-making suggestions for the efficient implementation of MTSS(B)'.
\section{Methodology- CANDI-QA}
\label{sec:Methodology}

The overall methodology is designed to systematically evaluate the performance of different prompting strategies for large language models (LLMs) on a set of specialized behavioral health questions. Our pipeline, as summarized in Figure~\ref{fig:bottomfig}, begins with the curation of a proprietary dataset by human experts from the School Behavioral Health Academy (SBHA). This dataset comprises 315 questions of two distinct types: T1 and T2. We then apply three different prompting strategies (P1, P2, and P3) to a chosen language model to generate responses for all questions. The evaluation process diverges for the two question types. For T1 questions, we leverage a set of gold-standard answers provided by our expert coaches to assess the quality of the model-generated responses using a suite of reference-based metrics, including Cosine Similarity, BERTScore, MoverScore, and BLEURT. Conversely, for T2 questions, which lack gold-standard answers, we employ a retrieval-augmented generation (RAG) evaluation framework based on the RAGAs library. In this framework, we utilize responses generated by the Qwen model as our retrieved context and evaluate the LLM-generated responses against them using metrics such as context precision, response relevance, faithfulness, and semantic similarity. This dual-pronged approach allows for a comprehensive and robust evaluation of the LLM's performance across different question types and prompting strategies.

\begin{figure}[t]
\centering
\includegraphics[width=\columnwidth]{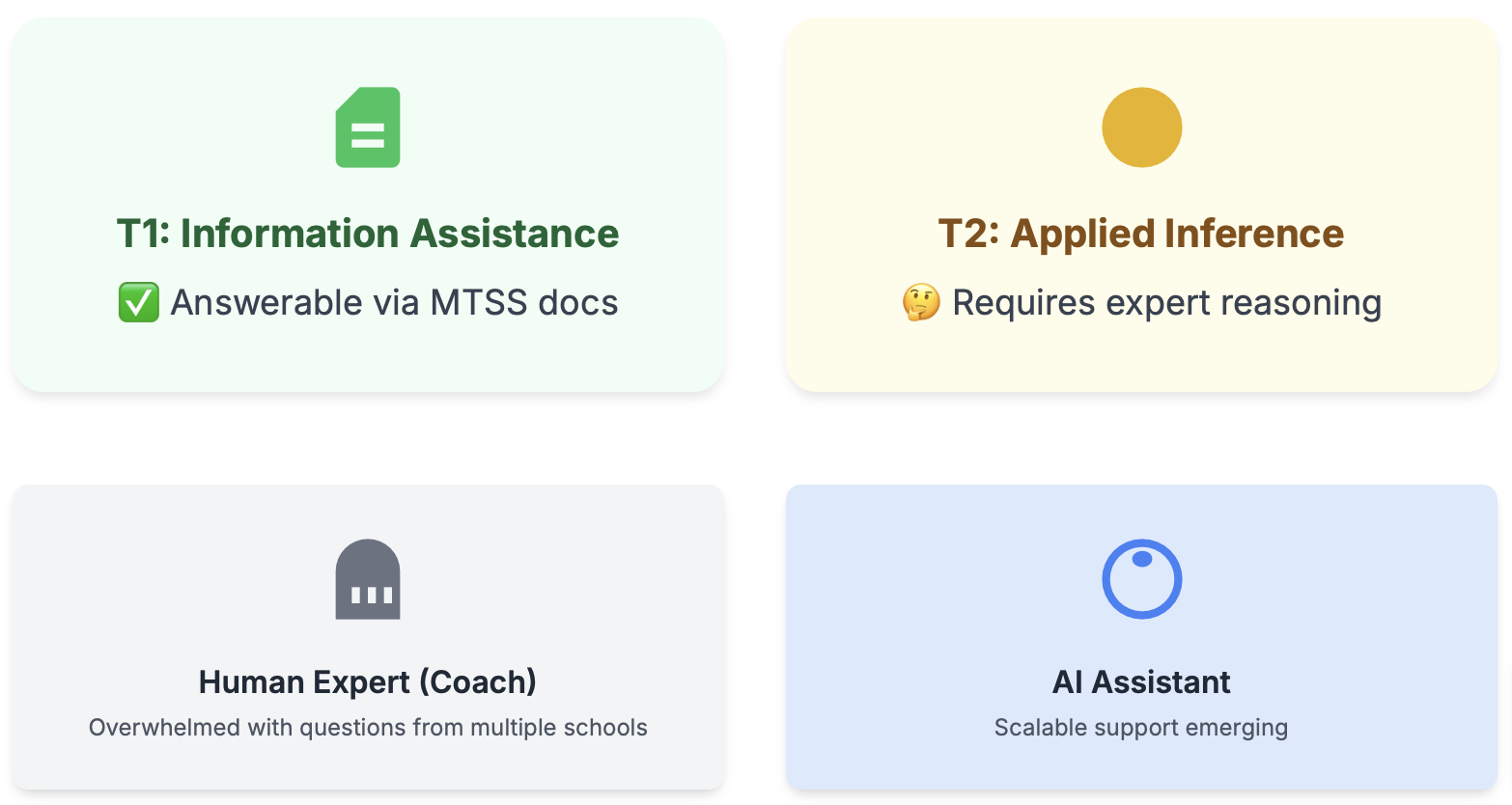}
\caption{Features of CANDI-QA}
\label{fig:bottomfig}
\end{figure}

\subsection{Dataset Curation}

The dataset for this study was meticulously curated with the assistance of human experts from the School Behavioral Health Academy (SBHA). The dataset consists of a total of 315 questions, categorized into two types: T1 and T2. T1 questions are designed to test foundational knowledge and are characterized by a set of 90 questions. T2 questions, comprising 225 questions, are more complex and require a deeper understanding of contextual nuances. The questions were developed to reflect the specific challenges and inquiries faced by MTSS (Multi-Tiered System of Support) teams, with the goal of providing a robust testbed for the LLMs in a real-world application setting.

\subsection{Prompting Strategies}

To assess the impact of different prompting techniques on model performance, we employed three distinct prompting strategies, labeled P1, P2, and P3. Each of these strategies was applied to generate responses from the chosen language model for all 315 questions in the dataset. The specific details and design of each prompting strategy are outlined below:
\begin{itemize}
    \item \textbf{P1: No-Context Prompting} \\
    In this setup, the model receives only the user query. No additional contextual or role-based information is provided.
    
    \item \textbf{P2: Role-Aware Prompting} \\
    The system prompt includes contextual information describing the user’s role and a corresponding role description. This setup provides the model with role-specific background to guide its responses.
    
    \item \textbf{P3: Task-Aware Prompting} \\
    The system prompt includes a high-level objective of the AI assistant, along with the user's role and role description. This strategy offers the model a comprehensive context that situates the question within the broader task it is meant to support.
\end{itemize}

\subsection{Evaluation of T1 Questions}

For the 90 T1 questions, we conducted a reference-based evaluation. Our human experts, who serve as coaches for the MTSS team, provided ``gold-standard'' answers for each question. The language model responses for each of the three prompting strategies (P1, P2, and P3) were then evaluated against these gold-standard answers using a variety of metrics designed to capture different aspects of semantic and syntactic similarity. The metrics used for this evaluation were:

\begin{itemize}
    \item \textbf{Cosine Similarity}: Measures the cosine of the angle between two vectors, representing the semantic similarity between the model response and the gold answer.
    \item \textbf{BERTScore}: A metric that leverages contextual embeddings from BERT to calculate the similarity between two sentences.
    \item \textbf{MoverScore}: A distance-based metric that measures the ``work'' required to transform one sentence into another, providing a measure of semantic distance.
    \item \textbf{BLEURT}: A learned metric that is trained to predict human judgments of text quality, offering a more human-aligned evaluation.
\end{itemize}

\subsection{Evaluation of T2 Questions}

For the 225 T2 questions, which are more open-ended and do not have a single gold-standard answer, we adopted a retrieval-augmented generation (RAG) evaluation framework using the RAGAs library. This framework is designed to assess the quality of generated responses by measuring the alignment between the generated answer, the retrieved context, and the original question. The evaluation was structured as follows:

\paragraph{Context Generation}  
To create a benchmark for evaluation, we used the Qwen model to generate responses for the T2 questions, which served as the ``retrieved contexts'' for our analysis.

\paragraph{Response Evaluation}  
We evaluated the responses generated by our primary language model (using prompting strategies P1 and P3) against these retrieved contexts. The evaluation was performed using the following RAGAs metrics:

\begin{itemize}
    \item \textbf{Context Precision:} This metric assesses whether all of the retrieved contexts are relevant to the question. A higher score indicates that the retrieved information is more focused and useful for answering the query.
    \item \textbf{Response Relevancy:} This metric measures how relevant the generated response is to the original user input. It is calculated by using an LLM to generate a set of questions from the response and then computing the average cosine similarity between these generated questions and the original question. A high score means the response is a direct and complete answer, while a low score indicates it may be incomplete or contain redundant information.
    \item \textbf{Faithfulness:} This metric evaluates the factual consistency of the generated response with the retrieved context. It works by identifying all the claims made in the response and verifying whether each claim can be directly inferred from the provided context. A score of 1.0 indicates that all claims are supported by the context, and a score of 0.0 means none are.
    \item \textbf{Answer Semantic Similarity:} This metric evaluates the semantic similarity between the generated response and a reference answer. In our case, since we do not have a gold-standard answer, we used the Qwen-generated response as the reference to compare the semantic meaning of the two outputs. This evaluation uses a cross-encoder model to compute the similarity score, which ranges from 0 to 1, with higher scores indicating better alignment.
\end{itemize}

The evaluation was performed by treating the language model's response as \texttt{response} and the Qwen model's response as \texttt{retrieved\_contexts}. The evaluation was thus conducted on pairs of (response, retrieved\_contexts), where \texttt{response=<language model response on P1 or P3>} and \texttt{retrieved\_contexts=[<qwen response on P1 or P3>]}.

\section{Experiment Setup}
\label{sec:ExperimentalSetup}

\begin{table*}[t]
\renewcommand{\arraystretch}{0.95}
\centering
\scriptsize
\caption{Evaluation of Models on T1: Information Assistance Across Three Prompt Strategies}
\label{tab:t1-evaluation}
\resizebox{\textwidth}{!}{%
\begin{tabular}{|l|c|c|c|c||c|c|c|c||c|c|c|c|}
\hline
\multirow{2}{*}{\textbf{Model}} & \multicolumn{4}{c|}{\textbf{P1}} & \multicolumn{4}{c|}{\textbf{P2}} & \multicolumn{4}{c|}{\textbf{P3}} \\ \cline{2-13} 
 & \textbf{Cos-Sim} & \textbf{BERTScore} & \textbf{MoverScore} & \textbf{BLEURT} & \textbf{Cos-Sim} & \textbf{BERTScore} & \textbf{MoverScore} & \textbf{BLEURT} & \textbf{Cos-Sim} & \textbf{BERTScore} & \textbf{MoverScore} & \textbf{BLEURT} \\
\hline
Phi-4-mini-instruct (3.8B) & 0.3252 & 0.8087 & 0.6915 & -1.2195 & 0.3328 & 0.8105 & 0.6915 & -1.2116 & 0.3277 & 0.8098 & 0.6919 & -1.2301 \\
\hline
Deepseek R1 & 0.3155 & 0.8085 & 0.6904 & -1.1894 & 0.3355 & 0.8106 & 0.692	& -1.2144 & 0.3267 & 0.81 & 0.6903 & -1.2004 \\
\hline
Qwen3 (32B) & 0.4302 & 0.8206 & 0.7072 & -1.0651 & 0.4332 & 0.8223 & 0.7098	& -1.0794 & 0.4344 & 0.824 & 0.7089 & -1.0866 \\
\hline
Gemma 2 - 9B & 0.421 & 0.8199 & 0.6964 & -0.9172 & 0.4356 & 0.8219 & 0.6998 & -0.9326 & 0.4292 & 0.8231 & 0.6982 & -0.9762 \\
\hline
Mistral 7B Instruct & 0.3324 & 0.8097 & 0.6842 & -0.9252 & 0.3411 & 0.8109 & 0.6847 & -0.9564 & 0.3321 & 0.8122 & 0.682 & -0.9621 \\
\hline
LLaMA 3.1 (8B) - Instruct  & 0.3181 & 0.8086 & 0.6194 & -1.2324 & 0.338 & 0.8105 & 0.6917 & -1.2043 & 0.3311 & 0.8103 & 0.6943 & -1.1846 \\
\hline
Llama 3.2 1B instruct & 0.3091 & 0.8083 & 0.6894 & -1.2345 & 0.3367 & 0.8099 & 0.6931 & -1.2092 & 0.3299 & 0.8106 & 0.6926 & -1.2143 \\
\hline
Llama 4 maverick 7B	& 0.4199 & 0.8206 & 0.7022 & -0.7795 & 0.4391 & 0.822 & 0.6969 & -0.8742 & 0.4314 & 0.8238 & 0.6976 & -0.8976 \\
\hline
Llama 4 scout 17B & 0.4257 & 0.8202 & 0.6934 & -0.8143 & 0.4369 & 0.8218 & 0.6969 & -0.8744 & 0.4392 & 0.8224 & 0.6962 & -0.8685 \\
\hline

\end{tabular}
}
\end{table*}

\subsection{Datasets}
We introduce our novel dataset \textbf{CANDI-QA}, curated specifically for question answering tasks in the MTSS domain with the help of our MTSS expert coaches. The dataset consists of two main types of questions: T1-type questions (information assistance questions) and T2-type questions (applied inference type questions). The dataset consists of $90$ T1-type questions and $225$ T2-type questions, along with gold answers, provided by our experts, for each question. Each question has a corresponding role.

The MTSS framework consists of the following roles: Admin, MTSS Coach, Counsellor, Clinician, Psychologist, Nurse, and Teacher.

For T1-type questions, the dataset consists of the role-specific questions. These are the types of questions that could be asked by members from various roles to the coaches.

For T2-type questions, the dataset consists of case-based questions, where each case belongs to one of the three MTSS intervention tiers (tier 1, 2 or 3).

\subsection{Input Setup}
We experiment on the question-answering task using three different types of prompts. We empirically found that:

\begin{itemize}
    \item \textbf{P1-type prompt}: No context, only the question is passed to the model.
    \item \textbf{P2-type prompt}: Role-aware context (\{role\} + \{role description\}) and the question are passed to the model.
    \item \textbf{P3-type prompt}: Task-aware context (\{objective\} + \{role\} + \{role description\}) and the question are passed to the model.
\end{itemize}

\subsection{Metrics}
\label{subsec:metrics}

We use two distinct sets of evaluation metrics aligned with the nature of T1 and T2 questions.

\paragraph{T1 Metrics: Reference-Based Evaluation}  
T1-type questions are direct, factual queries and include gold-standard reference answers. We apply the following reference-based metrics:

\begin{itemize}
    \item \textbf{Cosine Similarity:} Given two embedding vectors $\vec{a}$ and $\vec{b}$ for the model response and the gold answer, respectively,
    \[
    \text{CosSim}(\vec{a}, \vec{b}) = \frac{\vec{a} \cdot \vec{b}}{\|\vec{a}\| \|\vec{b}\|}
    \]
    
    \item \textbf{BERTScore:} For reference tokens $r_i$ and candidate tokens $c_j$, BERTScore computes:
    \[
    \text{BERTScore} = \frac{1}{|C|} \sum_{c \in C} \max_{r \in R} \text{sim}(c, r)
    \]
    where $\text{sim}(c, r)$ is the cosine similarity between contextual embeddings of token $c$ and token $r$.

    \item \textbf{MoverScore:} Measures the minimal cost of transforming one sentence into another based on token embeddings and their distances:
    \[
    \text{MoverScore}(X, Y) = \min_{T \in \mathcal{T}} \sum_{i,j} T_{i,j} \cdot d(e_i, e_j)
    \]
    where $T$ is a transport matrix and $d(e_i, e_j)$ is the distance between embeddings $e_i$ and $e_j$.

    \item \textbf{BLEURT:} Uses a pre-trained regression model $f_\theta(x, y)$ to compute a score:
    \[
    \text{BLEURT}(x, y) = f_\theta(x, y)
    \]
    where $x$ is the model response and $y$ is the gold answer.
\end{itemize}

\paragraph{T2 Metrics: Reference-Free Evaluation (RAGAs Framework)}  
T2-type questions are open-ended and evaluated without gold references, using retrieval-augmented context instead. Let $q$ be the original question, $r$ the generated response, and $c$ the retrieved context (from Qwen).

\begin{itemize}
    \item \textbf{Context Precision:}
    \[
    \text{Precision}_{context} = \frac{|c \cap \text{Rel}(q)|}{|c|}
    \]
    where $\text{Rel}(q)$ denotes the set of relevant context units to the query.

    \item \textbf{Response Relevancy:}
    \[
    \text{Rel}(q, r) = \cos(\phi(q), \phi(r))
    \]
    where $\phi(\cdot)$ is a sentence embedding function.

    \item \textbf{Faithfulness:}
    \[
    \text{Faith}(r, c) = \frac{|\text{Claims}(r) \cap \text{Supported}(c)|}{|\text{Claims}(r)|}
    \]
    where $\text{Claims}(r)$ are atomic factual assertions and $\text{Supported}(c)$ are those verifiable by the context.

    \item \textbf{Semantic Similarity:}
    \[
    \text{Sim}_{sem}(r, c) = f_{\text{cross-enc}}(r, c)
    \]
    where $f_{\text{cross-enc}}$ is a cross-encoder model that computes the similarity score between response and context.
\end{itemize}

\subsection{Implementation Details}
We evaluate a combination of open-source and proprietary language models to benchmark performance across diverse model sizes and architectures. All open-source models are accessed via HuggingFace using their official identifiers, ensuring reproducibility and public accessibility. The full list of models used in our experiments includes:

\paragraph{Model Identifiers}  
We evaluate the following models in our experiments. Open-source models are accessed from HuggingFace, while proprietary models are accessed via the Groq API.

\begin{itemize}
    \item \texttt{microsoft/Phi-4-mini-instruct-3.8B}
    \item \texttt{deepseek-ai/DeepSeek-R1-Distill-Llama-8B}
    \item \texttt{qwen/qwen3-32b} 
    \item \texttt{google/gemini-pro-1.5}
    \item \texttt{google/gemma-2-9b-it}
    \item \texttt{meta-llama/Meta-Llama-3-8B-Instruct}
    \item \texttt{meta-llama/Meta-Llama-3-1B-Instruct}
    \item \texttt{meta-llama/Meta-Llama-4-Maverick-7B} 
    \item \texttt{meta-llama/Meta-Llama-4-Scout-17B} 
    \item \texttt{mistralai/Mistral-7B-Instruct-v0.2}
\end{itemize}

\paragraph{Prompt Settings}  
Each model is evaluated using three prompting strategies: P1 (no context), P2 (role-aware context), and P3 (task-aware context). Prompts are constructed by combining system-level instructions (e.g., role descriptions and objectives) with user queries depending on the prompt type.

\paragraph{Generation Settings}  
For all open-source models, the following generation hyperparameters are used to maintain consistency across all model generations:
\begin{itemize}
    \item \texttt{max\_new\_tokens}: 512
    \item \texttt{temperature}: 0.6
    \item \texttt{top\_p}: 0.9
    \item \texttt{do\_sample}: \texttt{True}
    \item \texttt{batch\_size}: 16
\end{itemize}


\paragraph{Data Settings}  
The curated CANDI-QA dataset contains 315 total questions, divided into 90 T1-type and 225 T2-type questions. Each question is passed through all three prompt variants for every model. For T1 questions, the generated outputs are compared to gold-standard answers using reference-based metrics. For T2 questions, the outputs are evaluated using RAGAs metrics with Qwen model responses serving as the retrieval context.

The full codebase, model identifiers, and evaluation scripts will be released upon publication to support reproducibility.
\section{Baseline Solution - MTSS-Net}
\label{sec:mtssnet}

\begin{table*}[t]
\centering
\caption{Evaluation of Models on T2: Applied Inference Across Two Prompt Strategies (P1 and P3)}
\label{tab:t2-evaluation}
\resizebox{\textwidth}{!}{%
\begin{tabular}{|l|c|c|c|c||c|c|c|c|}
\hline
\multirow{2}{*}{\textbf{Model}} & 
\multicolumn{4}{c||}{\textbf{P1}} & 
\multicolumn{4}{c|}{\textbf{P3}} \\
\cline{2-9}
& \textbf{Context precision} & \textbf{Response relevancy} & \textbf{Faithfulness} & \textbf{Semantic similarity} 
& \textbf{Context precision} & \textbf{Response relevancy} & \textbf{Faithfulness} & \textbf{Semantic similarity} \\
\hline
Phi-4-mini-instruct (3.8B) & 0.6612 & 0.3696 & 0.7442 & 0.8044 & 0.7215 & 0.3724 & 0.8055 & 0.8101 \\
\hline
LLaMA 3.1 (8B)             & 0.6415 & 0.3542 & 0.7512 & 0.8182 &0.7124 & 0.3653 &0.7856 &0.8236 \\
\hline
Mistral 7B Instruct          &0.6733 & 0.3661 & 0.7565 & 0.8077 & 0.7231 & 0.4005 & 0.8014 & 0.8336 \\
\hline
\end{tabular}
}
\end{table*}

To support future research on contextual alignment, we introduce a neuro-symbolic baseline framework, \textbf{MTSS-Net}, designed for the CANDI-QA benchmark. MTSS-Net models how structured, role-aware reasoning can enhance language model (LM) performance in specialized domains. It includes two variants: \textbf{v1}, which leverages structured JSON-based representations for context tracking, and \textbf{v2}, which builds on v1 by adding source attribution and role conditioning to enable more transparent and grounded reasoning.

MTSS-Net is used to address both task types in the CANDI-QA setting:

\begin{itemize}
    \item \textbf{T1:} Information Assistance Questions – direct, factual queries that benefit from structured lookup.
    \item \textbf{T2:} Applied Inference Questions – scenario-specific questions requiring multi-hop reasoning or role-sensitive interpretation.
\end{itemize}

\subsection{MTSS-Net v1: Structured Context Representation}

MTSS-Net v1 encodes each QA instance as a structured JSON object capturing four elements:

\begin{itemize}
    \item Task type (T1 or T2)
    \item Role or identity of the user (e.g., teacher, counselor)
    \item Intervention tier (Tier 1, 2, or 3)
    \item Relevant policy or procedural references
\end{itemize}

This structure acts as a semantic scaffold, supporting symbolic rule-based reasoning or enabling the construction of more interpretable prompts for small LMs. JSON is chosen for its clarity, interpretability, and ease of use with transformer-based models. This variant is especially effective for T1 tasks, where answers often rely on retrieving clearly defined information based on structured inputs.

\subsection{MTSS-Net v2: Role-Aware Reasoning with Source Attribution}

MTSS-Net v2 enhances the structured representation of v1 with three additional components:

\begin{itemize}
    \item \textbf{Curated QA pairs} linked to expert-verified or synthesized documents,
    \item \textbf{Role conditioning} to align responses with the expected responsibilities of different users (e.g., administrator vs. paraprofessional),
    \item \textbf{Source citation} that explicitly grounds responses in the underlying documentation or procedures they rely on.
\end{itemize}

These enhancements aim to handle the complexity of T2 questions, where understanding the user’s intent, their institutional role, and the scenario-specific context is critical. Role-aware conditioning helps the system tailor responses in ways that match real-world team dynamics. The source attribution feature also promotes explainability, allowing users to trace answers back to supporting evidence.

Together, MTSS-Net v1 and v2 illustrate persistent challenges in aligning LMs with real-world domain expectations, especially around context disambiguation and inference—issues often overlooked by traditional alignment benchmarks. As part of this work, we release MTSS-Net as a reproducible baseline to support future research into interpretable and role-sensitive QA systems.



\section{Results}
\label{sec:Results}

\subsection{Evaluation of T1 Questions: Information Assistance}
This evaluation assesses how effectively language models answer T1 (information-assistance) questions in the behavioral health domain using three prompting strategies: P1 (zero-shot), P2 (role-aware), and P3 (context-rich). Table~\ref{tab:t1-evaluation} summarizes performance across Cosine Similarity, BERTScore, MoverScore, and BLEURT. \textbf{Qwen3 (32B)} consistently outperforms others, with the highest semantic alignment (Cos-Sim: 0.4302–0.4344). Smaller models like \textbf{Llama 3.2 1B instruct} perform lowest. Role-aware prompting (P2) yields modest improvements over P1, showing the value of incorporating role context. However, gains from P2 to P3 are marginal, suggesting that task-level context provides limited benefit for straightforward factual queries.

\subsection{T2 Evaluation Summary}
Table~\ref{tab:t2-evaluation} shows that all models benefit from task-aware prompting (P3) over P1. \textbf{Mistral 7B Instruct} performs best, with highest faithfulness (0.8014) and semantic similarity (0.8336). \textbf{Phi-4-mini-instruct} and \textbf{LLaMA 3.1 (8B)} also show clear gains in context precision and faithfulness. These results suggest that T2 tasks require more than retrieval—they benefit from structured, role- and task-aware prompting to support grounded, scenario-sensitive reasoning.

\section{Discussions}
\label{sec:Discussions}


Our experiments highlight several challenges in applying LLMs to real-world, domain-specific tasks such as MTSS QA. Strong performance on open-domain benchmarks does not guarantee alignment with expert preferences in niche domains, suggesting that preference alignment is not domain-generalizable.

Models benefit significantly from contextual disambiguation. Prompting strategies that include role and task-aware context (P2 and P3) yield improved results, especially for T2 questions requiring inferential reasoning. However, prompting alone is insufficient—scenario-specific queries demand structured guidance beyond plain text.

Structured input using formats like JSON offers clear advantages. LLMs show better semantic precision and interpretability when given predictable, serialized inputs, likely due to easier token-level parsing and alignment.

In the MTSS domain, role context and task type are critical. T1 questions depend on factual lookup, while T2 questions demand procedural reasoning and situational awareness. These findings support augmenting QA datasets with structured context and metadata to better reflect real-world usage.

This points toward hybrid approaches that combine symbolic scaffolding with neural reasoning, especially in enterprise and educational settings.

Future work may explore self-reflective or planning-based modules to improve multi-hop reasoning, along with task-type-aware benchmarking to better differentiate between factual and inferential QA tasks.

\section{Conclusion}
\label{sec:Conclusion}

This work highlights the importance of contextual alignment for domain-specific QA, particularly within the MTSS behavioral health framework. Through systematic evaluations across prompting strategies and model types, we demonstrate that role- and task-aware prompts significantly improve performance over zero-shot baselines.

A key limitation of this study is its reliance on static prompting without adaptive reasoning or feedback mechanisms. Future work will explore integrating self-reflective or planning-based modules, as well as expanding the dataset to cover broader MTSS scenarios.

The MTSS-Net neuro-symbolic baseline solution, along with supplementary materials, will be released to support further research and reproducibility.

\newpage



\bigskip

\bibliography{main}

\end{document}